\documentclass[runningheads]{llncs}

\usepackage[T1]{fontenc}
\def\doi#1{\href{https://doi.org/\detokenize{#1}}{\url{https://doi.org/\detokenize{#1}}}}
\usepackage{graphicx}
\usepackage{amsmath}
\usepackage{adjustbox}
\usepackage{listings}
\begin{document}
\title{Detecting early signs of depression in the conversational domain: The role of transfer learning in low-resource scenarios}
\titlerunning{Detecting early signs of depression in the conversational domain}
%
\author{Petr Lorenc\inst{1} \and
Ana-Sabina Uban\inst{2,3} \and
Paolo Rosso\inst{2}
\and
Jan Šedivý\inst{4}}
 \authorrunning{P. Lorenc et al.}
\institute{Faculty of Electrical Engineering , Czech Technical University in Prague, Czechia \\ \email{lorenpe2@fel.cvut.cz} \and
PRHLT research center, Universitat Politècnica de València, Spain
\\ \email{prosso@dsic.upv.es}
\and
Faculty of Mathematics and Computer Science, University of Bucharest, Romania
\\ \email{ana.uban+prof@gmail.com}
\and
CIIRC, Czech Technical University, Czechia \\ \email{jan.sedivy@cvut.cz}}
\maketitle              
\begin{abstract}
The high prevalence of depression in society has given rise to the need for new digital tools to assist in its early detection. To this end, existing research has mainly focused on detecting depression in the domain of social media, where there is a sufficient amount of data. However, with the rise of conversational agents like Siri or Alexa, the conversational domain is becoming more critical. Unfortunately, there is a lack of data in the conversational domain. We perform a study focusing on domain adaptation from social media to the conversational domain. Our approach mainly exploits the linguistic information preserved in the vector representation of text. We describe transfer learning techniques to classify users who suffer from early signs of depression with high recall. We achieve state-of-the-art results on a commonly used conversational dataset, and we highlight how the method can easily be used in conversational agents. We publicly release all source code\footnote{https://github.com/petrLorenc/mental-health}.

\keywords{Depression detection  \and Conversational domain \and Transfer learning}
\end{abstract}
\section{Introduction}
The World Health Organization\footnote{https://apps.who.int/iris/handle/10665/254610} estimates that over 300 million people suffer from depression. However, there are approximately only 70 mental health professionals available for every 100,000 people in high-income nations, and this number can drop to 2 for every 100,000 in low-income countries \cite{6_Lee2019CaringFV}. It leads to a high percentage of the world population suffering from depression, and only a tiny fraction has access to psychiatric care to detect early signs of any mental illness, including depression. On the other hand, almost everyone has access to smartphones \cite{40_internet_use}, and with the rise of conversational agents like Siri or Alexa \cite{10_pichl2020alquist}, people are getting used to communicating with their smartphones or smart speakers on daily basis \cite{9_Gabriel2020FurtherAI}. This evolution of conversational agents allows for building mental health applications, like the virtual therapists TalkToPoppy!\footnote{https://www.talktopoppy.com/}. It brings new challenges to recognize early signs of mental illnesses such as depression immediately during the conversation.

Unfortunately, there is a scarcity of conversational data usable for the detection of early signs of depression. The lack of these data is due to several problems. Authors of such datasets need to collect a representative sample of data to balance positive and especially negative examples \cite{16_balancing_data}, and typically cross-reference data with medical records, but this process can raise ethical issues. Some of these problems are mitigated in social media, such as Reddit or Twitter. On social media platforms, we usually get access to vast amounts of self-labeled data \cite{17_davide2016atest}. In addition, self-stated diagnoses remove some overheads of annotating data but increase the false-positive noise. Therefore, based on the data scarcity, we focus on sequential transfer learning \cite{31_phang2019sentence} which helps mainly in the target domain with limited data and it is based on transferring knowledge from a related domain with a sufficient amount of data,. 

As shown in \cite{8_mental_disorders}, there are several indicative symptoms of depression like a loss of interest in everyday activities, feelings of worthlessness, and also a change in the use of language \cite{1_xezonaki2020affective,2_han_depression}. Because the change in language use can be detected, there are several lexicon-based approaches \cite{33_liwc,34_nrc_mohammad2013crowdsourcing} to extracting semantic features from text. These approaches suffer from a limited size of vocabulary and require human annotation. This paper investigates sentence embeddings \cite{13_cer2018universal,14_devlin2019bert} and whether they can capture these changes in the use of language without the need for a time-consuming design of lexicons . Furthermore, recent works focus on attention mechanisms \cite{32_yang-etal-2016-hierarchical}, showing promising results in the social media \cite{18_UBAN2021480}. We propose a novel model that combines attention mechanisms with sentence embeddings.

Our contributions can be summarized as follows:
\begin{enumerate}
    \item We evaluate the usage of sentence embeddings to detect change in the use of language for the detection of early signs of depression and its possible combination with attention mechanisms;
    \item We explore sequential transfer learning from social media to the conversational domain;
    \item We achieve state-of-the-art results on retrieving indications of early signs of depression in the conversational domain;
\end{enumerate}

The rest of the paper is organized as follows. Section \ref{sec:metholodogy} introduces the used data, methodology for transfer learning, and our novel model. Section \ref{sec:experimental} introduces the experimental setting and Section \ref{sec:ablations} shows the results and proposes possible usage.

\section{Related work}

Previous works on textual depression detection were mainly focused on detecting early signs of depression in the social media domain \cite{17_davide2016atest,yates-etal-2017-depression,wolohan-etal-2018-detecting,coppersmith-etal-2015-clpsych}, in contrast to the conversational domain, which has received rarer attention \cite{4_gratch-etal-2014-distress}.

\subsection{Early Sign of Depression Detection}

\textbf{Datasets} - The most remarkable conversational dataset related to virtual mental health applications is The Distress Analysis Interview Corpus (DAIC) \cite{4_gratch-etal-2014-distress}. As for datasets focusing on online forums and social networks, there is a sufficient number of them. The eRisk dataset \cite{17_davide2016atest}, RSDD dataset \cite{yates-etal-2017-depression} or the dataset introduced by \cite{wolohan-etal-2018-detecting} are extracted from Reddit. Additionally, there is the CLPsych 2015 Shared Task \cite{coppersmith-etal-2015-clpsych} focusing on depression and PTSD on Twitter.

\textbf{Depression Detection in the Conversational Domain} - Several works \cite{1_xezonaki2020affective,2_han_depression,Dinkel2019TextbasedDD} have focused on the DAIC-WOZ dataset \cite{4_gratch-etal-2014-distress}. In \cite{2_han_depression}, the authors used the Hierarchical Attention Model \cite{32_yang-etal-2016-hierarchical} with a low-level representation of words based on word vector representation --- GloVe \cite{44_glove} embeddings. They used word embeddings together with different types of the Hierarchical Attention Model to obtain a high-level representation of participant texts. Similarly, \cite{1_xezonaki2020affective} uses Hierarchical Attention Model with a combination of lexical features like LIWC \cite{33_liwc}, NRC Emotion Lexicon (Emolex) \cite{34_nrc_mohammad2013crowdsourcing}, and many others. In contrast to \cite{1_xezonaki2020affective,2_han_depression}, we perform an extensive study of transfer learning techniques, and similarly to \cite{Dinkel2019TextbasedDD}, we find that proper hyperparameters are critical for training the model.

\textbf{Depression Detection in Social Media} - The primary studies focusing on depression detection in the social media domain are also based on lexicon-based features, as well as word embeddings \cite{18_UBAN2021480}. Similarly to \cite{1_xezonaki2020affective,2_han_depression},  \cite{18_UBAN2021480} uses the Hierarchical Attention Model with additional features. Nevertheless, in \cite{3_depression_in_social}, it is shown that simpler models with additional emotional and linguistic features can achieve comparable results. Because of that, we include a simple baseline model, such as Logistic regression \cite{41_logistic_regression}.

\subsection{Transfer learning}

Transfer learning is used to improve a learner from one domain by transferring information from a related domain \cite{19_Weiss2016ASO}. It has been shown that if the two domains are related, transfer learning can potentially improve the results of the target learner \cite{transfer_learning}.
Furthermore, it was shown by \cite{transfer_learning} that transfer learning can improve the performance in anxiety and depression classification. They examine the performance of deep language models for pre-training the model. Similarly, he authors of \cite{abed2019transfer_learning_2} demonstrate usefulness of transfer learning for depression detection from social media postings, applied to the eRisk \cite{17_davide2016atest} dataset. In \cite{uban2022transfer}, the authors show that transfer learning can be effective for improving prediction performance for disorders where little annotated data is available. They explore different transfer learning strategies for both cross-disorder (across disorders) and cross-platform transfer (across different social media platforms).


\section{Methodology}
\label{sec:metholodogy}
In the following section, we introduce the datasets and metrics. We also discuss our novel chunk-based model.

\subsection{Data}

To study detection of depression, we set The Distress Analysis Interview Corpus as our target dataset. Source dataset was eRisk data which is labeled for early signs of depression, where for each user with depression there are possibly texts posted before the diagnosis or the onset of the disease. Additionally, we use General Psychotherapy Corpus, leaving other datasets for future research. All datasets contain two categories of participant, depressed (positive) and non-depressed (negative). Each participant is linked with a sequence of sentences. The data statistics for all mentioned datasets, can be seen in Table \ref{tab:data-statistics} and the description of these datasets follows.

The Distress Analysis Interview Corpus - Wizard-of-Oz (DAIC-WOZ) is part of a larger corpus, the Distress Analysis Interview Corpus (DAIC) \cite{4_gratch-etal-2014-distress}. These interviews were collected as part of a more significant effort to create a conversational agent that interviews people and identifies verbal and non-verbal indicators of mental illness \cite{38_inproceedings}. Original data collected include audio and video recordings and extensive questionnaire responses. DAIC-WOZ includes the Wizard-of-Oz interviews, conducted by a virtual interviewer called Ellie, controlled by a human therapist in another room. The data have been transcribed and annotated for various verbal and non-verbal features. 

The eRisk dataset \cite{17_davide2016atest} consists of data for the Early Detection of Signs of Depression Task presented at CLEF (specifically 2017\footnote{https://clef2017.clef-initiative.eu/}). The texts were extracted from the social media platform Reddit, and it uses the format described in \cite{17_davide2016atest}.




Additionally, we apply our models to the General Psychotherapy Corpus (GPC) collected by the “Alexander Street Press”\footnote{Can be found at https://alexanderstreet.com/} (ASP). This dataset contains over 4,000 transcribed therapy sessions, covering various clinical approaches and mental health issues. The data collection was compiled according to \cite{1_xezonaki2020affective} and we chose only transcripts related to depression. It results in 147 sessions. Additionally, we randomly chose 201 sessions annotated with mental illnesses different than depression in order to use them as a control group. 



\begin{table}[h!]

\caption{Data statistics. \textit{A} stands for all data. \textit{P} stands for utterances of participant.}
\label{tab:data-statistics}
\centering
\setlength{\tabcolsep}{4pt}

\begin{tabular}{l|c|c|c|c|c|c}
\hline
Dataset         &                 & \# dialogues & \# utterances & vocabulary & labels 0/1 & train/valid/test \\ \hline
DAIC-WOZ        & A              & 189         & 20,857       & 8,272      & 133/56   & 107/35/47        \\ \hline
DAIC-WOZ        & P & 189         & 10,505       & 8,263      & 133/56  & 107/35/47        \\ \hline \hline
eRisk           & A & 1304         & 811,586      & 322,634    & 214/1090        & 387/97/820 \\ \hline \hline
GPC             & A              & 348         & 54,588       & 54,844     & 201/147  & 208/70/70        \\ \hline
GPC             & P & 348         & 26,860       & 45,205     & 201/147  & 208/70/70        \\ \hline
\end{tabular}

\end{table}

\subsection{Metrics}

As suggested by \cite{1_xezonaki2020affective,2_han_depression}, we used the Unweighted Average Recall (UAR) between the ground-truth and the predicted labels associated with each participant (see Equation \ref{eq:uar}). As shown in \cite{1_xezonaki2020affective}, the UAR metric is also suitable when the label distribution of the dataset is unbalanced. Additionally, we measure Unweighted Average Precision (UAP), same as UAR but recall is substituted by precision and macro F1 score (macro-F1) for completeness. 


\begin{equation}\label{eq:uar}
        UAR = \frac{Recall_0 + Recall_1}{2} = \frac{\frac{TP_0}{TP_0+FN_0} + \frac{TP_1}{TP_1+FN_1}}{2}
\end{equation}

where $TP_0,FP_0,FN_0$ are true positives, false positives and false negatives for non-depressed participants, respectively. $TP_1,FP_1,FN_1$ represent true positives, false positives and false negatives for depressed participants, respectively.

\subsection{Chunk-based classification}
\label{seq:chunk}

\begin{figure}
    \centering
    \includegraphics[width=0.9\columnwidth]{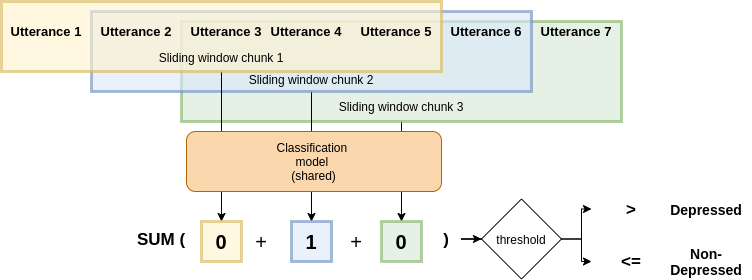}
    \caption{The conversation is divided into sliding window chunks. Each chunk is classified independently from the others. The positive/negative labels ratio is used to determine the best threshold.}
    \label{fig:slide-window}
\end{figure}

Since natural conversation is an infinite sequence of utterances, our proposed Chunk-based model works based on a sliding window as shown in Figure \ref{fig:slide-window}. We classify each chunk of the conversation with a binary label, then we sum up all the obtained classifications (zeros for chunks corresponding to non-depressed participants and ones for depressed participants). Then we divide the sum by the number of chunks to normalize for different conversation lengths. To obtain a prediction for an entire conversation, we use a threshold on the ratio between positive and negative labels. More concretely, the conversation $C_i$ is created during an iterative process of conversation. Then, each conversation $C$ is composed of a set of utterances $U$, where each utterance $u_i$ is composed of one or more sentences $S$, as shown in Equation \ref{eq:conversation}. 

\begin{equation}\label{eq:conversation}
    C_i = \{U_i\} = \{u_1, u_2, ..., u_n\} = \{\{S_1\}, \{S_2\}, ... \{S_n\}\}
\end{equation}

Further, in order to allow for the iterative evaluation of the conversation in a real-time setting, we performed classification using a sliding window at the chunk level. Each chunk $X_i$ is labeled according to the label $y$ of the conversation $C_i$ from which the chunk is derived. These chunks are overlapping, as shown in Figure \ref{fig:slide-window}. All chunks of the same length (shown for the length of three in Equation \ref{eq:slide}) have same label $y_i$ derived from label $y_i$ of conversation $C_i$.

\begin{equation}
\begin{split}\label{eq:slide}
    X_1  &= (u_1, u_2, u_3) \\
    X_2  &= (u_2, u_3, u_4) \\
    &       ... \\
    X_N  &= (u_\text{N-2}, u_\text{N-1}, u_\text{N})
\end{split}
\end{equation}

Firstly, we train the model $M$ to classify each chunk of the conversation $M(X_n)$ into a binary label $y$. After training, we classify all $N$ chunks obtained in the validation set data and perform the search for the best threshold $T_\text{best}$ for distinguishing conversations of depressed participants from non-depressed ones. The best value of the threshold $T_\text{best}$ is based on the accuracy over the whole validation set. The expression describing a classification $f$ for a conversation $C$ is shown in Equation \ref{eq:clasificaion}. 

\begin{equation}\label{eq:clasificaion}
\begin{adjustbox}{width=0.9\textwidth}

$f(C) = f(\{X_1, X_2, ..., X_N\}) = \left\{ \begin{array}{cc} 
                \text{depressed} & \hspace{5mm} \frac{\sum_N z(X_n, 0)}{\sum_N z(X_n, 1)} > T_\text{best} \\ \\
                \text{non-depressed} & \hspace{5mm} \frac{\sum_N z(X_n, 0)}{\sum_N z(X_n, 1)} <= T_\text{best} 
                \end{array} \right.$
\end{adjustbox}
\end{equation}            
where
\begin{equation}
z(X_n, i) = \left\{ \begin{array}{cc} 
                1 & \hspace{5mm} M(X_n) = i \\ \\
                0 & \hspace{5mm} \text{otherwise} 
                \end{array} \right.
\end{equation}

Obviously, a smaller chunk size allows us to make more precise predictions gradually as new participant utterances occur. However, a smaller chunk size leads to loss of context information for particular classification.

\section{Experimental setting}
\label{sec:experimental}

This section highlights the setting of the suggested models and discusses their usability. We measure the performance of our model over the DAIC-WOZ dataset. First setting was without transfer learning. Then we measure the influence of transfer learning as a way to improve the performance of our model on the DAIC-WOZ dataset using the eRisk and GPC datasets as source domains.

In our experiments, we use recurrent neural networks as model $M$ for each chunk, concretely Long Short Term Memory (LSTM) \cite{15_lstm} which is commonly used for sequence labeling in the conversational domain \cite{10_pichl2020alquist}. As input to the LSTM model, we included multi-head self-attention Transformer architecture \cite{14_devlin2019bert,reimers-2019-sentence-bert} and Deep Average Architecture \cite{13_cer2018universal}. Specifically, the input to the model $M$ consists of sentence embeddings obtained from the pooled output of the fine-tuned BERT \cite{reimers-2019-sentence-bert} (sBERT), the output of the Universal Sentence Encoder - Deep Average Network \cite{13_cer2018universal} (DAN), or the output of the Universal Sentence Encoder - Transformer based \cite{13_cer2018universal} ($\text{USE}_5$). Additionally, as in \cite{1_xezonaki2020affective,2_han_depression,18_UBAN2021480}, we evaluate an attention mechanism. We use two settings, the Hierarchical Attention Network (HAN) based on the GloVe embedding as in \cite{18_UBAN2021480} and pure attention based on a dot-product of the hidden states of LSTM and learned attention weights.



We follow the evaluation process described in \cite{2_han_depression}. However, in contrast to \cite{1_xezonaki2020affective,2_han_depression}, we perform Bayesian hyperparameter optimization \cite{21_frazier2018tutorial_bayes}. The reported results are with the best performing setting of hyperparameters. 

We follow the common setup of transfer learning using the fine-tuning approach in a cross-domain setting \cite{transfer_learning}. More specifically, we train the model on source domain data until convergence. The learning rate is then reduced in order to avoid catastrophic forgetting \cite{39_MCCLOSKEY1989109}. Then, the training continues in the target domain. The decrease of the learning rate also reduces the overwriting of useful pretrained information and maximizes positive transfer. The weights of sentence embedding model are frozen.

\section{Results and Ablation Experiments}
\label{sec:ablations}

To demonstrate the importance of the vocabulary size, we focus on the performance of the logistic regression model using different sizes of the vocabulary. Therefore, we include logistic regression \cite{41_logistic_regression} over bag-of-words vectors \cite{42_bag_of_words} as the baseline model. We extracted several types of vocabulary based on the utterances of participants in the DAIC-WOZ (\textbf{3k} - 3000 words), based on the utterances of participants and therapists in the DAIC-WOZ (\textbf{6k} - 6000 words), as well as based on posts in the eRisk dataset (\textbf{20k} - 20000 words). The best results were achieved with the logistic regression and the \textit{3k} vocabulary consisting of 3000 most used words based only on the participants' utterances - UAR (0.583), UAP (0.603) and macro-F1 (0.593), in contrast to UAR (0.579), UAP (0.561) and macro-F1 (0.570) for the \textit{6k} vocabulary or UAR (0.583), UAP (0.580) and macro-F1 (0.581) for the \textit{20k} vocabulary. We infer that a more extensive vocabulary probably introduces additional noise for the classification model. We use the best performing vocabulary for the rest of the experiments when working with logistic regression. 



To confirm the difference between the vocabulary of depressed and non-depressed patients, we then examine the weights $W_x$ learned by the logistic regression model. We look at the most significant logistic regression weights in absolute value, both positive and negative, and map the weights to the vocabulary words. The results indicate that words like environment (-7.5), open-minded (-6.3), or accomplish (-4.7) correspond with a non-depressed patient. In contrast, insignificant (5.36), television (5.66), or pollution (+6.1) relate to a depressed patient. It confirms results reported in \cite{1_xezonaki2020affective,2_han_depression}, showing the difference between the depressed and non-depressed groups in terms of the use of language, more specifically, at the level of word usage.



Finally, we test the performance of the classification models proposed in Section \ref{seq:chunk}. Our results, shown in Table \ref{tab:results}, show the high performance of transfer learning approach. We achieve a new state-of-the-art result, specifically using the chunk-based model based on bidirectional LSTM over sentence embedding. The input to the models was based on the Universal Sentence Encoder - Transformer ($\text{USE}_5$).

The transfer learning achieved a notable outcome on data from the domain of social media. We also assume that this was caused by the implicit ability of sentence embedding to capture different language characteristics \cite{30_liu2019linguistic}. We discuss these results in more depth in Section \ref{sec:ablations}. According to our results, attention is not beneficial for chunk-based classification. We assume it is caused by the sliding window chunk-based classification, where the attention mechanism is not fully utilized. 

\begin{table}[h!]
\caption{Results - \textbf{LR} stands for Logistic Regression, \textbf{HAN} - Hierarchical Attention Model,  \textbf{Chunk-biLSTM} - our Chunk-based model based on bidirectional LSTM, \textbf{DAN} - sentence embeddings based on Deep Average Network, $\textbf{USE}_5$ - sentence embeddings based on Transformer trained by \cite{13_cer2018universal}, \textbf{sBERT} - sentence embeddings based on Transformer trained by \cite{reimers-2019-sentence-bert}, \textbf{att} stands for the attention mechanism.}
\setlength{\tabcolsep}{4pt}
\label{tab:results}
\centering
\begin{adjustbox}{width=1\textwidth}
\begin{tabular}{l|c|c|c}
\hline
              \textbf{Model}    & \multicolumn{3}{c}{\textbf{Unweighted Average Recall}}                                                                                                                                      \\ \hline
HCAN \cite{2_han_depression}         & \multicolumn{3}{c}{0.54}                                                                                                                                                           \\ \hline
HLGAN \cite{2_han_depression}        & \multicolumn{3}{c}{0.60}                                                                                                                                                           \\ \hline
HAN \cite{1_xezonaki2020affective}          & \multicolumn{3}{c}{0.54}                                                                                                                                                           \\ \hline
HAN + L \cite{1_xezonaki2020affective}      & \multicolumn{3}{c}{\textbf{0.72}}                                                                                                                                                           \\ \hline
                  & \textbf{DAIC-WOZ} & \begin{tabular}[c]{@{}c@{}}\textbf{eRisk/GPC}\\ \textbf{without fine-tuning}\end{tabular} & \begin{tabular}[l]{@{}c@{}}\textbf{eRisk/GPC}\\ \textbf{with fine-tuning}\end{tabular} \\ \hline
\textbf{LR + unigrams 3k}  & 0.553          & 0.559 / 0.547 & 0.613. / 0.553         \\ \hline
\textbf{HAN + GloVe}       & 0.541          & 0.511 / 0.535 & 0.529 / 0.613          \\ \hline
\textbf{Chunk-biLSTM} + \textbf{DAN}     & 0.595          & 0.559 / 0.470 & 0.625 / 0.541          \\ \hline
\textbf{Chunk-biLSTM} + \textbf{DAN} + \textbf{att} & \textbf{0.666} & 0.630 / 0.333 & 0.676 /  0.494         \\ \hline
\textbf{Chunk-biLSTM} + $\textbf{USE}_5$     & 0.660          & \textbf{0.651} / 0.440 & \textbf{0.803} / \textbf{0.690} \\ \hline
\textbf{Chunk-biLSTM} + $\textbf{USE}_5$ + \textbf{att} & 0.529          & 0.541 / \textbf{0.589} & 0.613 / 0.595          \\ \hline
\textbf{Chunk-biLSTM} + \textbf{sBERT}     & 0.440          & 0.505 / 0.523 & 0.613 / 0.541          \\ \hline
\textbf{Chunk-biLSTM} + \textbf{sBERT} + \textbf{att} & 0.442          & 0.5 / 0.523   & 0.636 / 0.577          \\ \hline
\end{tabular}
\end{adjustbox}

\end{table}

Also, we present various ablation experiments to provide some interpretations of our findings.

\textbf{Are sentence embeddings able to encode information present in lexicon-based features?} We are interested in verifying whether lexicon-based features are helpful to our classifiers or if the sentence embeddings already encode the information provided by the lexicons. To test this assumption, we performed another experiment using the best-performing architecture, where we added linguistic characteristics (emotions and LIWC) as input features along with sentence embeddings. Results, shown in Table \ref{tab:results2}, indicate that there is no improvement in using linguistic characteristics. Therefore, we conclude that the sentence embeddings already include linguistic characteristics needed for detection.

\begin{table}[h!]
\setlength{\tabcolsep}{3pt}
\centering
\caption{Results - \textbf{Chunk-biLSTM} - our Chunk-based model based on bidirectional LSTM, $\textbf{USE}_5$ - sentence embeddings based on Transformer trained by \cite{13_cer2018universal}, \textbf{feat} stands for additional features.}
\label{tab:results2}
\begin{adjustbox}{width=1\textwidth}
\begin{tabular}{c|c|c|c}
\hline
                  & DAIC-WOZ & \begin{tabular}[c]{@{}c@{}}eRisk/GPC \\ without fine-tuning\end{tabular} & \begin{tabular}[c]{@{}c@{}}eRisk/GPC\\ with fine-tuning\end{tabular} \\ \hline
\textbf{Chunk-biLSTM} + $\textbf{USE}_5$     & \textbf{ 0.660}    & \textbf{ 0.651} / 0.440 & \textbf{0.803} / 0.690      \\ \hline
\textbf{Chunk-biLSTM} + $\textbf{USE}_5$ + \textbf{feat} & 0.565    & 0.541 / 0.410  & 0.597 / 0.511          \\ \hline
\end{tabular}
\end{adjustbox}
\end{table}

\textbf{Is the size of the source domain dataset more critical than domain relatedness?} Results in Table \ref{tab:results} show that transfer learning can help with improving the classification performance on the conversational dataset. At the same time, we find a surprising result showing that using GPC as the source domain underperforms the setting in which eRisk data is used as the source domain, even though GPC is a more similar type of data to our target dataset (they are both conversational datasets). We assume, that the smaller data size can cause poor performance when using the GPC dataset: to test this hypothesis, we evaluate a smaller version of eRisk (eRisk small). With 66,516 utterances and 388/96/820 participants. The new smaller dataset is closer to the GPC dataset in respect to size. Results, in Table \ref{tab:results3}, suggest that the size of the source domain dataset is as much important as domain closeness.

\begin{table}[h!]
\setlength{\tabcolsep}{3pt}
\centering
\caption{Results - \textbf{Chunk-biLSTM} - our Chunk-based model based on bidirectional LSTM. The double horizontal line divides the table with results on \textbf{eRisk} (above) and results on \textbf{eRisk-small} (below).}
\label{tab:results3}
\begin{tabular}{c|c|c|c}
\hline
                  & DAIC-WOZ & without fine-tuning & with fine-tuning \\ \hline
\textbf{Chunk-biLSTM} + $\textbf{USE}_5$   & \textbf{0.660}    &\textbf{ 0.651}    & \textbf{0.803}  \\ \hline \hline
\textbf{Chunk-biLSTM} + $\textbf{USE}_5$     & \textbf{0.660}   & 0.642    & 0.690 \\ \hline
\end{tabular}

\end{table}

\subsection{Usage}

Because our approach is based on sliding window chunks, we are able to perform a real-time evaluation of the conversation as soon as the number of utterances reaches the size of the sliding window chunk. Our experiments are performed with 50 as the chunk size. This allows for including the classification model as another part of the Natural Language Understanding (NLU) unit commonly used in conversational agents \cite{10_pichl2020alquist,11_finch2020emora}. As opposed to other models proposed in literature such as \cite{1_xezonaki2020affective,2_han_depression,18_UBAN2021480}, our suggested model is independent of external lexical features, such as lexicon-based features, and therefore it can be run in parallel with other NLU units.

\section{Conclusion and Future Work}
\label{sec:conclusion}
In this paper, we addressed the problem of detecting early signs of depression in the conversational domain. We achieve state-of-the-art results on the DAIC-WOZ dataset using transfer learning from the social media domain. The proposed model was based on a sequence of chunk classification and uses a recurrent neural network with sentence embedding as input features. Additionally, we show that the attention mechanism is not beneficial for our chunk-based model. We show that transfer learning helps improve the performance in a domain with a lack of data utilizing data from a related domain. Additionally, we demonstrate that the size of the source dataset is as important as the domain relatedness between the source and the target. We also suggest a possible usage of our model as a tool for therapists, who may retrieve early signs of depression from a broad range of conversational systems.



\subsection{Ethical concern}

This paper showed a possible usage of automatic techniques for detecting early signs of depression. Unfortunately, false positive and false negative cases can cause tremendous damage when used in the conversational agent. We claim that if our model or proposed techniques are used in real-life scenarios, they has to be supervised by a qualified therapist. Also, as we have shown, domain adaptation plays a crucial role too. Our approach, even if quite general, has to be carefully adapted to a specific domain. We also suggest not relying on the classification system only, but using it as another source of information for a qualified therapist. With this setting, we can minimize possible harm and allow the therapist to speed up their work.

\subsubsection{Acknowledgments.}

The research work of Petr Lorenc and Jan Šedivý was partially supported by the Grant Agency of the Czech Technical University in Prague, grant
(SGS22/082/OHK3/1T/37). The research work of Paolo Rosso was partially funded by the Generalitat Valenciana under DeepPattern (PROMETEO/2019/121). The work of Ana Sabina Uban was carried out at the PRHLT Research Center during her postdoc internship. Her work was also partially funded by a grant from Innovation Norway, project "Virtual simulated platform for automated coaching-testing", Ref No 2021/331382.




%
%
%
\bibliographystyle{splncs04}
\bibliography{mybibliography}
%




\end{document}